# When Does Causality Help in Multivariate Time-Series Anomaly Detection?

Long Zhang*
China Satellite Maritime Tracking and Control Department
Jiangyin, China, 214400
zlinux@foxmail.com

Wei Huang
China Satellite Maritime Tracking and Control Department
Jiangyin, China, 214400
weeihuang@163.com

Aishui Rao
China Satellite Maritime Tracking and Control Department
Jiangyin, China, 214400
56585972@qq.com

Jian Hu
China Satellite Maritime Tracking and Control Department
Jiangyin, China, 214400
mayfan11@163.com

Binxiang Wu
China Satellite Maritime Tracking and Control Department
Jiangyin, China, 214400
3221276638@qq.com

Jie Li
China Satellite Maritime Tracking and Control Department
Jiangyin, China, 214400
JieLi63547@qq.com

Chen Gong
China Satellite Maritime Tracking and Control Department
Jiangyin, China, 214400
2577048174@qq.com

Linxi Feng
China Satellite Maritime Tracking and Control Department
Jiangyin, China, 214400
assky@163.com

**Abstract—Causality-aware scoring is common in multivariate time-series anomaly detection, yet the literature rarely states when it helps. Holding one causality-guided autoencoder (DeCauAE) fixed, we vary graph quality, anomaly type and protocol against ground truth. Graph quality fixes the payoff: on a twenty-variable DAG the gain over a graph-free autoencoder decays as the graph is corrupted. Anomaly type fixes the target: the structural factor lifts window F1 from 0.800 to 0.976 where anomalies break conditional structure, and its advantage vanishes below a break severity $s \approx 0.3$. Protocol fixes the ranking: point-adjusted F1 lets a random score match the full model, while range-aware VUS separates the autoencoder family from the classical baselines. On real benchmarks the causal module pays only where density makes a system eligible and cross-subsample reproducibility selects entities inside it (+0.022 vs −0.029 on SMAP, $p = 0.031$); enabling it everywhere costs 0.012 VUS-ROC. The structural factor also transfers to classical detectors (+0.05–0.06 VUS-ROC).**

## I. Introduction

Multivariate time-series anomaly detection has accumulated a long roster of deep architectures [1], and a recent strand adds causal structure to the scoring rule: a graph over variables is estimated from training data and used to reweight, condition or discount per-channel reconstruction [2]–[5]. An anomaly that breaks the relation between a variable and its parents is invisible to per-channel statistics, yet the evidence for the added machinery is thin and the protocols in common use are themselves criticised [6], [7].

This paper asks directly: when does causality help? Holding one causality-guided autoencoder fixed, we vary three factors against ground truth: graph quality (a conditional Granger test under Benjamini–Hochberg control, the true graph corrupted from $\rho = 0$ to $\rho = 1$), anomaly type (breaks of conditional structure that preserve marginals versus marginal-visible spikes), and protocol (point-adjusted F1, raw F1, AUROC and the range-aware VUS family [8], on three public benchmarks and two controlled ones).

The protocol, not the model, decides which story is visible: on the controlled benchmarks the causal machinery produces large, reproducible gains that follow the graph-quality gradient, while on MSL, SMAP and PSM it changes point-adjusted F1 by less than entity-to-entity variation and a random-score control is indistinguishable from the full model. The deliverable is not a stronger detector but a two-stage gate that says when to switch the causal module on.

## II. Related Work

Deep multivariate detectors model channels jointly through reconstruction and forecasting [1], [9], [10]; causal discovery supplies the structure causality-aware detectors build on [2]–[5], with CAAD [4] closest and classical Granger regression [11] the workhorse. MSL, SMAP [12] and PSM [13] are the standard benchmarks. Kim et al. [6] document point-adjustment inflation, VUS [8] proposes range-aware AUCs and TSB-AD [7] benchmarks protocols at scale; none varies the causal graph or the anomaly type, which our audit adds with the two-stage gate of Section IV-E.

## III. Method

A memory-distilled denoising autoencoder is trained on a decontaminated sample of the training series; a conditional-Granger graph drives a per-variable weighting and a structural-deviation score; attribution and drift adaptation reuse the same pieces.

Decontamination. We train the autoencoder on overlapping bags of the training window and drop rows whose median out-of-bag reconstruction error is implausible. On clean data it is a no-op, so it is part of the pipeline but not of the claim.

Causal structure. For every ordered pair (i, j) we fit the restricted and unrestricted Granger autoregressions [11] with a fixed lag order $L = 5$ (the true maximum lag of the synthetic systems; sensitivity in Section V):

$$x_{j,t} = c_j + \sum_{u=1}^{d} \sum_{\ell=1}^{L} \beta_{u\ell}^{(j)} x_{u,t-\ell} + \varepsilon_t^{(j)} \quad (1)$$

the unrestricted fit uses the lags of all d variables and the restricted fit removes the L lagged copies of i, and

$$F_{i\to j} = \frac{(\mathrm{SSR}_j^{\mathrm{red}} - \mathrm{SSR}_j^{\mathrm{full}})/L}{\mathrm{SSR}_j^{\mathrm{full}}/\nu} \quad (2)$$

an edge i→j is kept when the F statistic survives Benjamini–Hochberg control at q = 0.01 over all d(d−1) tests; the replacement for the thresholded lagged cross-correlation most causality-aware detectors use (Section IV-B).

Each variable then receives a reconstruction weight that decays with its median out-of-bag error, normalised by its interquartile range, so unreliable channels are down-weighted.

Scoring. The first factor is the 90th percentile of the weighted per-variable squared reconstruction error on a sliding window. The second is a structural deviation: for every variable j with a parent we fit, on normal data, a ridge linear structural equation from its lagged parents,

$$u_{j,t} = \frac{1}{\sigma_j}\left|x_{j,t} - \sum_{i\in \mathrm{pa}(j)} \sum_{\ell=1}^{L} w_{i\ell}^{(j)}\, x_{i,t-\ell}\right| \quad (3)$$

where σj is the training residual standard deviation and pa(j) the parent set; C(w) is the largest per-variable 90th percentile of the one-step-ahead residuals over the window. The two are fused as

$$s(w) = \alpha\, R(w)/R_{99}^{\mathrm{tr}} + (1-\alpha)\, C(w)/C_{99}^{\mathrm{tr}}, \quad \alpha = 0.7 \quad (4)$$

so magnitude-like anomalies are caught by R and structure-breaking ones by C (Section IV-B). Attribution discounts each variable's score by what it would inherit from above-threshold parents through the graph:

$$r_i = z_i - \beta \sum_{p\in \mathrm{pa}(i)} \max(0,\ z_p - \tau)\ \frac{G_{pi}}{\sum_q G_{qi}} \quad (5)$$

where β = 0.7 weights the explained evidence and the gate τ = 1.0 keeps quiet parents from eroding leaf faults. At test time, drift adaptation replaces each channel's raw score by a rolling z-score over W = 20 samples referenced to the fixed training standard deviation.

## IV. Experiments

### A. Setup

Controlled benchmarks. Two synthetic systems with ground-truth graphs: a chain with d = 5 variables (lag-2 edges) and a random DAG with d = 20 and lags 1–5. A-spikes are visible in the affected variable's marginal; B-breaks change only the conditional relation between a variable and its parents. Five seeds per configuration.

Benchmarks. MSL and SMAP [12] are per-channel telemetry streams consumed without feature selection: 55 MSL entities (27 with test anomalous segments) and 54 SMAP; PSM [13] is 25 channels with 87,841 labelled points in one entity. Detection uses 27 MSL and 54 SMAP entities, reproducibility 26 and 50 (long enough for half-sample splitting) and density all 81 with a graph estimate.

Protocol and baselines. The default metric is point-adjusted F1 with a threshold selected on test labels (oracle thr.); we also report raw F1, AUROC and the range-aware VUS family [8] with B = 100, using Wilcoxon signed-rank tests across entities with Holm correction. Baselines are PCA, Isolation Forest [14], LOF, spectral residual and a vanilla autoencoder [15] sharing the backbone, window, optimiser and threshold but with no graph or decontamination, plus a random-score control (i.i.d. U(0,1), 3 seeds). Ablations: −dual removes C(w), −causal uses the empty graph, −decont skips decontamination. USAD [9], TranAD [16], OmniAnomaly [17], GDN [10] and Anomaly Transformer [18] are not reproduced: our numbers speak to one autoencoder family.

### B. Graph quality and its payoff on the controlled benchmarks

The conditional Granger estimator recovers the chain exactly in four of five seeds and raises edge F1 on the twenty-variable DAG from 0.513 (thresholded cross-correlation) to 0.998 (structural Hamming distance 98.0 → 0.2). How much graph quality does detection need? Replacing a fraction of the true edges by spurious ones, detection degrades but never collapses (Fig. 1): on the DAG the full model retains 61 % of its gain even when every true edge is replaced, while a random graph of the same density stays flat at 0.863. The structural factor only needs parents to model residuals against, so an incorrect graph of the right density still provides them.

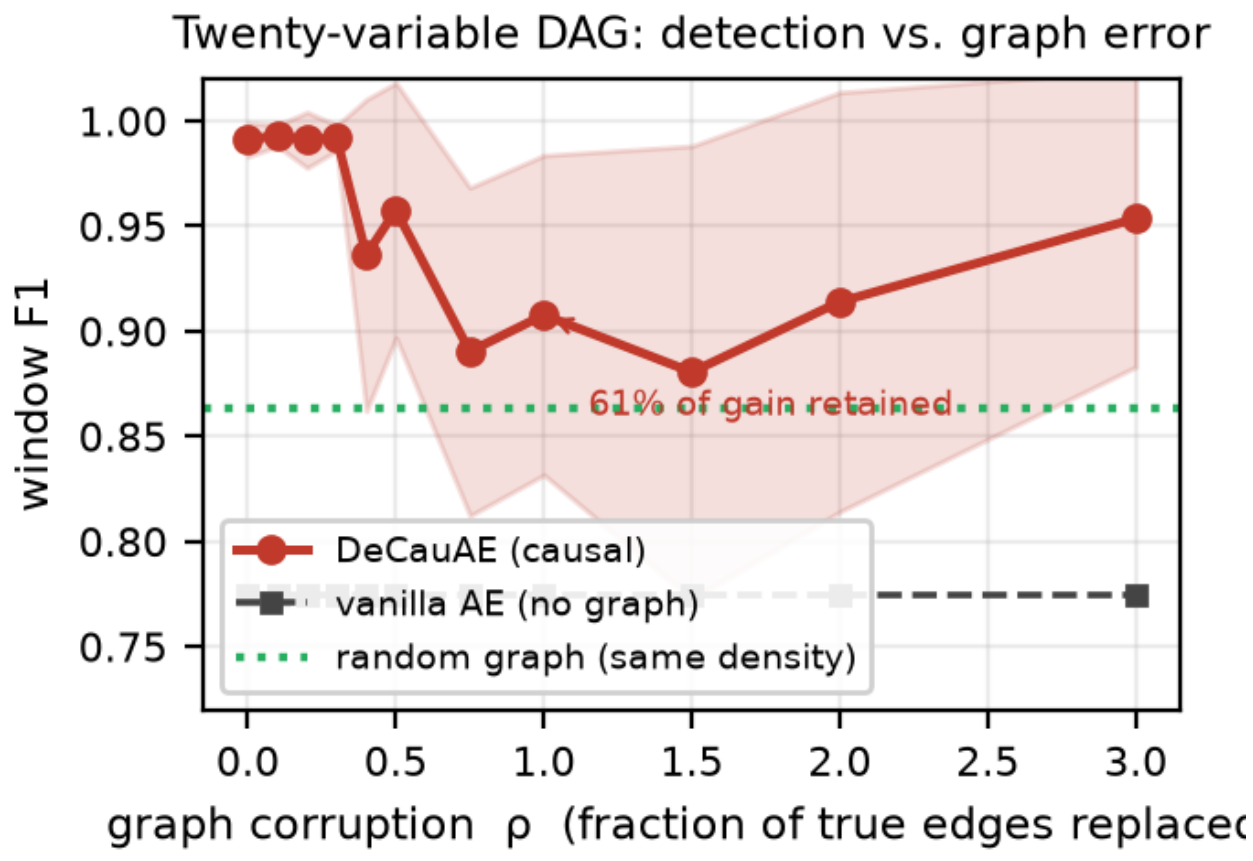

Fig. 1. Detection on the twenty-variable DAG as the estimated graph is progressively replaced by spurious edges (ρ = fraction of true edges replaced). DeCauAE degrades but never returns to the graph-free level, and a random graph of the same density (dotted) sits at the corrupted level, so density rather than edge identity carries the gain. Bands: ± 1 s.d. over five seeds.

Window-level detection follows the anomaly classes. On the chain the classical baselines reach F1 0.705 and the vanilla autoencoder 0.800 against 0.976 for the full model; removing the causal module returns it to 0.781 and removing decontamination leaves 0.969, so the gain comes almost entirely from the structural factor. On the DAG the full model reaches 0.993 against 0.775.

Graph reproducibility separates the worlds. On the controlled benchmarks every estimated edge reappears in 0.99 of ten half-length subsamples, but on PSM the estimator declares 486 of 600 possible edges significant (a near-complete graph) and the mean edge reappears in only 0.79 of subsamples, so per-segment DeCauAE and the vanilla autoencoder score identically (median ΔAUC +0.002 over 72 segments). The reconciliation is protocol-level (Section IV-D).

### *C. Detection on MSL, SMAP and PSM: the protocol decides the ranking*

TABLE I. POINT-ADJUSTED F1 (ORACLE THRESHOLD) ON THE THREE REAL BENCHMARKS, MACRO-AVERAGED OVER ENTITIES AND SEEDS; THE LAST TWO COLUMNS DROP POINT ADJUSTMENT AND THE THRESHOLD ON MSL. ON MSL NONE, INCLUDING THE THREE ABLATIONS, SURVIVES HOLM CORRECTION; ON SMAP TWO DO: SPECTRAL RESIDUAL ($\Delta$F1 = −0.150, HOLM $p < 0.001$) AND THE −DUAL ABLATION ($\Delta$F1 = +0.019, $p = 0.0002$); PSM'S SINGLE ENTITY EXCLUDES ENTITY-LEVEL TESTS.

| method | PA-F1 MSL | PA-F1 SMAP | PA-F1 PSM | raw F1 MSL | AUROC MSL |
|---|---|---|---|---|---|
| Random score (control) | 0.874 | 0.842 | 0.981 | 0.199 | 0.502 |
| PCA | 0.907 | 0.841 | 0.907 | 0.385 | 0.628 |
| Isolation Forest | 0.833 | 0.730 | 0.909 | 0.273 | 0.576 |
| LOF | 0.892 | 0.824 | 0.894 | 0.323 | 0.612 |
| Spectral residual | 0.914 | 0.948 | 0.982 | 0.245 | 0.577 |
| Vanilla AE | 0.868 | 0.799 | 0.948 | 0.357 | 0.703 |
| DeCauAE (full) | 0.857 | 0.799 | 0.929 | 0.351 | 0.678 |
| −dual (no structural factor) | 0.855 | 0.780 | 0.937 | 0.357 | 0.689 |
| −causal (empty graph) | 0.862 | 0.788 | 0.940 | 0.358 | 0.691 |
| −decont (no decontamination) | 0.862 | 0.799 | 0.936 | 0.352 | 0.686 |

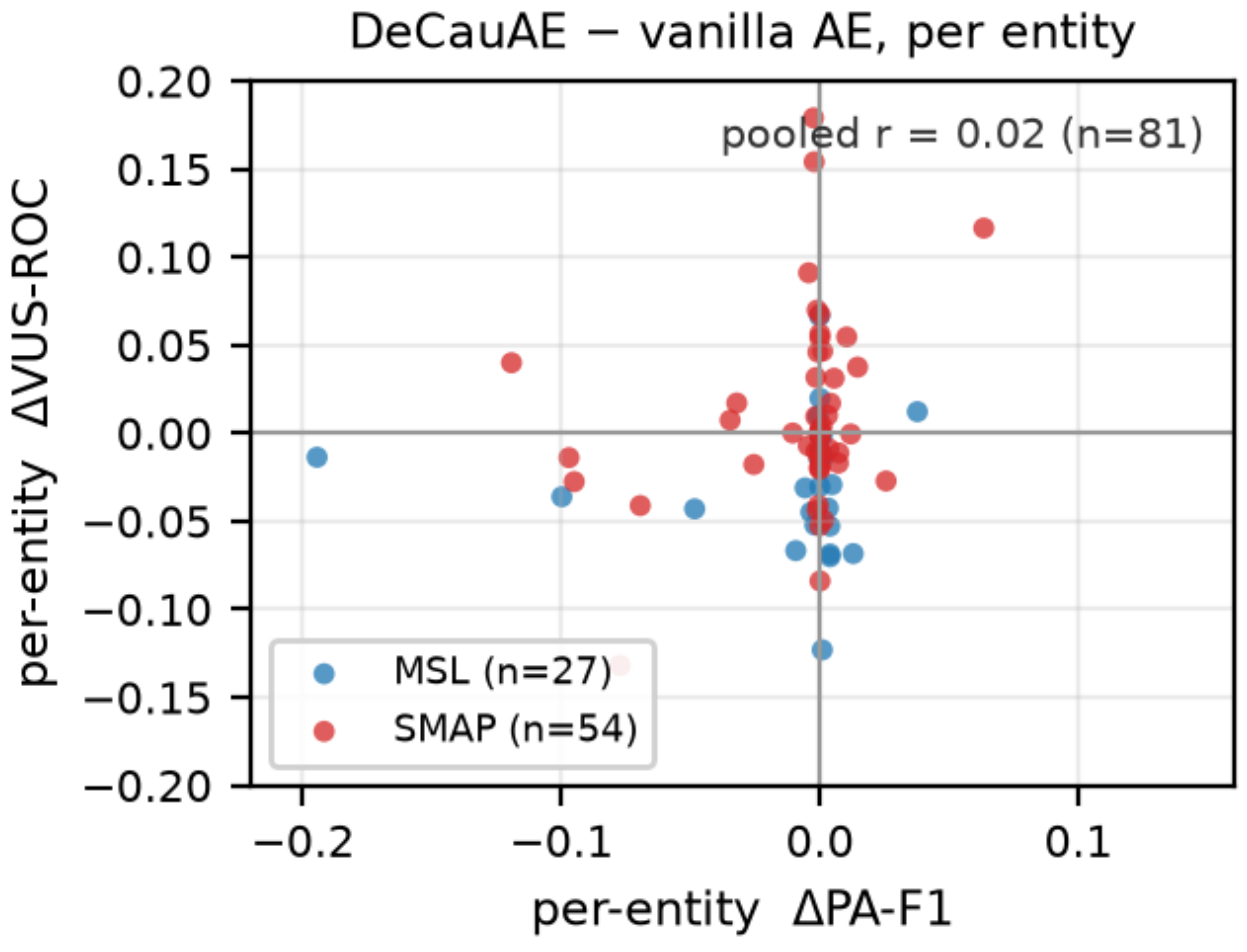

FIG. 2. PER-ENTITY AGREEMENT BETWEEN THE TWO PROTOCOLS ON THE REAL BENCHMARKS. FOR EACH OF THE 81 ENTITIES WE PLOT THE DECAUAE-MINUS-VANILLA GAIN UNDER POINT-ADJUSTED F1 AGAINST THE SAME GAIN UNDER VUS-ROC; THE POOLED CORRELATION IS $r = 0.02$, SO THE PROTOCOLS DO NOT RE-RANK A COMMON ORDERING BUT SELECT DIFFERENT ENTITIES. TWO POINTS LIE OUTSIDE THE AXES.

Three observations. First, under point adjustment the ranking is uninformative (Table I): the random control reaches 0.874 on MSL against 0.857 for the full model and is statistically indistinguishable from it ($p = 0.66$ across 27 entities; SMAP $p = 0.067$). Second, removing the adjustment collapses the absolute values (raw F1 0.351, AUROC 0.678 on MSL). Third, no MSL comparison survives Holm correction, while on SMAP two do. Per entity, the DeCauAE-minus-vanilla gain under point-adjusted F1 and its VUS-ROC counterpart correlate at $r = 0.02$ over the 81 real entities (Fig. 2), so the protocols select different entities rather than re-rank the same ones.

### *D. Range-aware and segment-level evaluation*

Point adjustment credits a prediction that overlaps a true event at all; the range-aware VUS family [8] removes that forgiveness by dilating ground-truth ranges by every b from 0 to B = 100 and averaging the AUC over b.

Two findings (Table II). The autoencoder family separates from the classical baselines far more cleanly than under point adjustment: on MSL the full model beats all four (Holm p between 0.028 and 0.0024). Inside the family the vanilla autoencoder reaches 0.679 ± 0.157 on MSL against 0.652 ± 0.158 (Holm $p = 0.0024$, 23 of 27 entities). An $\alpha$ sweep isolates the cause: raising the R/C weight raises macro VUS-ROC from 0.630 to 0.668, so the sparse MSL graphs (median density 0.006) leave parent sets too small for C to help. On SMAP the structural factor helps (+0.009, $p = 0.018$) while the causal weights cost accuracy (−0.013, $p < 0.001$), so the two autoencoders nearly tie (0.615 vs 0.619).

TABLE II. VUS-ROC AT B = 100 ON THE REAL BENCHMARKS (SEEDS AVERAGED PER ENTITY; POST HOC FROM THE DUMPED SCORES). UNDER THIS RANGE-AWARE METRIC THE AUTOENCODER FAMILY SEPARATES FROM THE CLASSICAL BASELINES, AND THE VANILLA AUTOENCODER SIGNIFICANTLY REVERSES THE FULL MODEL ON MSL.

| method | VUS-ROC MSL | VUS-ROC SMAP | VUS-ROC PSM |
|---|---|---|---|
| DeCauAE (full) | 0.652 | 0.615 | 0.753 |
| Vanilla AE | 0.679 | 0.619 | 0.749 |
| PCA | 0.585 | 0.599 | 0.753 |
| LOF | 0.589 | 0.582 | 0.701 |
| Isolation Forest | 0.545 | 0.549 | 0.665 |
| Spectral residual | 0.558 | 0.515 | 0.567 |

### *E. From graph reproducibility to a two-stage deployment gate*

Applying the subsample protocol to every training entity links the two halves of the paper. The rank correlation between reproducibility and the per-entity gain depends on the protocol: null against the point-adjusted gain (pooled $\rho = 0.041$, 76 entities), significant against the VUS-ROC gain (MSL $\rho = 0.406$, SMAP $\rho = 0.298$, pooled $\rho = 0.296$, $p = 0.0093$), and carried by the medium-reproducibility stratum ($\rho = +0.303$, $p = 0.031$, $n = 51$) rather than the low one ($\rho = -0.202$, n.s., $n = 23$).

The two graph-side quantities act at different levels. Table III recasts the gate as two stages: density decides whether a system is dense enough for conditional evidence to exist, and reproducibility then selects entities inside a system that passed it. Density correlates with the gain only across benchmarks (pooled $\rho = 0.245$, $p = 0.027$; SMAP median 0.068 against 0.006 on MSL); within a benchmark it is null (0.053 and 0.012).

Table III. A two-stage deployment gate and its accounting over the same 76 MSL+SMAP entities. "n on" is the number of entities the module would be enabled on; "ΔVUS (on)" the mean VUS-ROC gain of DeCauAE over the vanilla autoencoder among those; "rule gain" the same gain over all 76 entities (zero where the module is off); "% oracle" the fraction of the +0.0165 attainable by enabling the module exactly where its gain is positive.

| rule | n on | ΔVUS (on) | rule gain | % oracle |
|---|---|---|---|---|
| always on | 76 | −0.0118 | −0.0118 | — |
| never on (vanilla) | 0 | — | 0.0000 | — |
| oracle (only where +ve) | 28 | +0.0447 | +0.0165 | 100% |
| stage 1: density ≥ 0.056 | 38 | +0.0169 | +0.0085 | 51% |
| stage 1 + 2: repro ≥ 0.50 | 29 | +0.0230 | +0.0088 | 53% |
| stage 1 + 2: repro ≥ 0.589 | 22 | +0.0242 | +0.0070 | 43% |

Stage one separates the benchmarks: 38 of 50 SMAP entities clear the density threshold but none of the 26 MSL entities does, and on MSL no reproducibility threshold turns the gain positive (−0.025, −0.011, −0.007 at 0.5, 0.6, 0.7). Stage two works inside SMAP, splitting entities into +0.0217 (14 of 25 positive) and −0.0294 (Mann–Whitney $p = 0.031$ at the median 0.589; $p = 0.058$ at 0.60). Two caveats: the benchmarks differ an order of magnitude in density, so the first stage cannot be separated from benchmark identity; and the gate is not fitted to its own evaluation: leave-one-out medians reproduce the same 22 entities and +0.0070, and calibrating on MSL while applying to SMAP still gives +0.0221 against −0.0243 ($p = 0.058$).

The third precondition is met but not discriminative. The conditional-versus-marginal audit labels 89 % of MSL and 63 % of SMAP segments as conditional-structure breaks, so the anomalies the structural factor is built for are present, but the per-entity conditional fraction is degenerate as a selector (1.0 for every MSL entity and 67 % of SMAP ones), so no correlation with the gain can be estimated.

Deployability checklist. Before enabling the module: (i) align channels and estimate the graph on the training series alone; (ii) check density (eligibility); (iii) inside an eligible system, measure cross-subsample reproducibility against $\theta \approx 0.6$; (iv) confirm on known or injected anomalies that faults break conditional, not marginal, structure; (v) if all four pass, enable it and judge with a range-aware metric.

### F. Does the structural factor transfer beyond this autoencoder?

Every result so far fixes one autoencoder, so we ask whether the structural-deviation factor C carries signal of its own. We keep the graph, the ridge structural equations and C(w), discard the reconstruction factor R, and fuse C with a classical detector's z-normalised score as $\alpha \cdot \text{baseline} + (1 - \alpha) \cdot C$.

The causal factor is portable (Table IV). Adding C raises spectral residual by 0.047 on MSL and 0.061 on SMAP, and PCA by 0.061 on MSL and 0.012 on SMAP ($\alpha = 0.7$ gives the same picture), so the conditional-residual signal is real outside this autoencoder. The MSL reversal of Section IV-D is therefore not evidence that C is useless: fused with a detector that is not a reconstruction error, C helps on MSL too. What fails there is the particular R/C fusion of DeCauAE, not the structural factor.

Table IV. The structural-deviation factor C(w) plugged onto two classical baselines: VUS-ROC at B = 100, macro over entities (MSL 27, SMAP 54), one seed. Only the reconstruction factor R is replaced by the baseline's own score.

| method | VUS-ROC MSL | VUS-ROC SMAP |
|---|---|---|
| PCA | 0.578 | 0.592 |
| PCA + C ($\alpha = 0.5$) | 0.640 | 0.604 |
| Spectral residual | 0.558 | 0.515 |
| SR + C ($\alpha = 0.5$) | 0.604 | 0.577 |

### G. When does the causal gain appear?

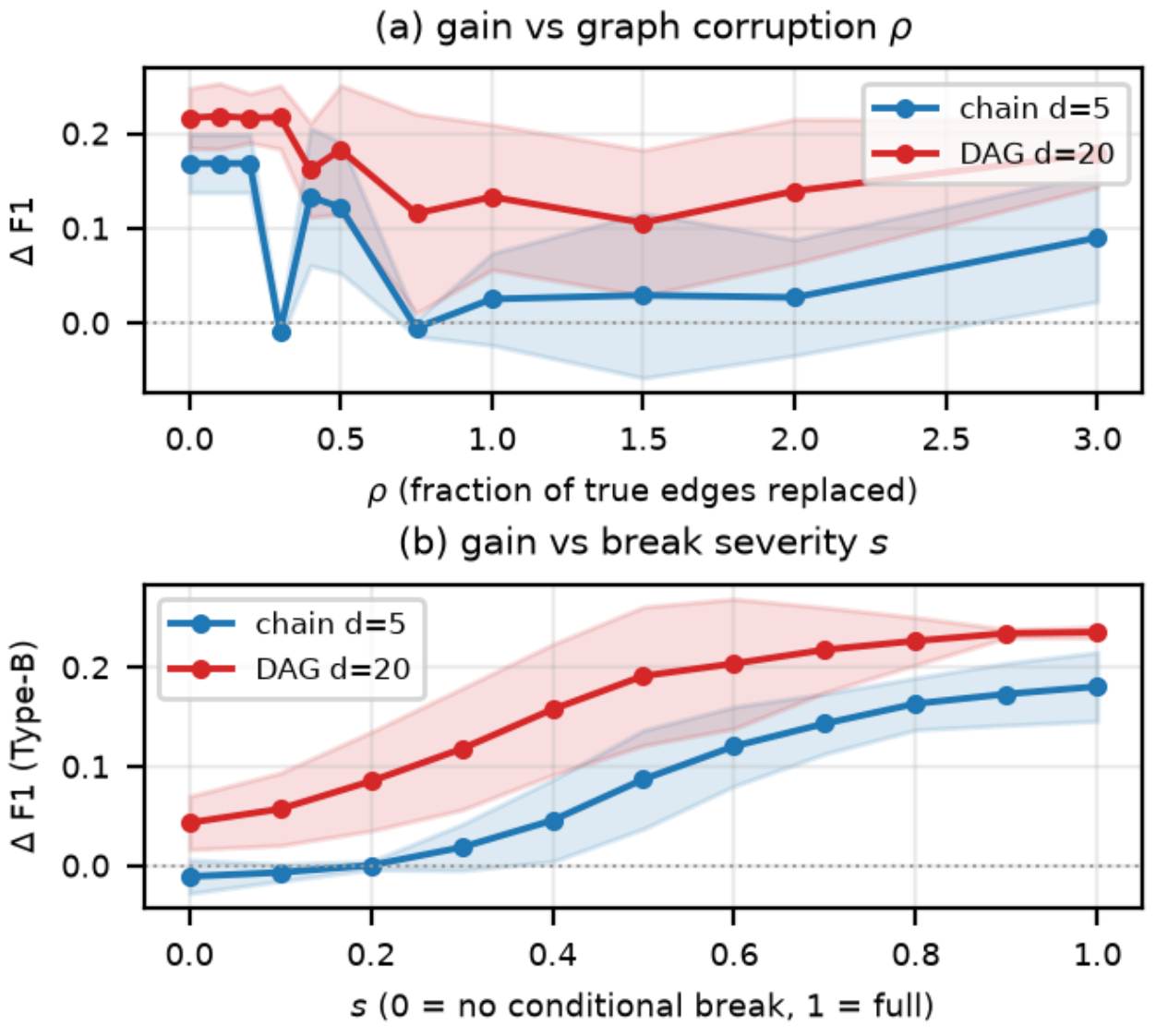

FIG. 3. THE CAUSAL GAIN IS BOUNDED FROM BOTH SIDES. (A) GAIN (DECAUAE − VANILLA) VERSUS GRAPH CORRUPTION P, RE-PLOTTED FROM THE SWEEP OF FIG. 1: IT DECAYS AS THE GRAPH DEGRADES. (B) GAIN ON TYPE-B WINDOWS VERSUS THE SEVERITY S OF THE CONDITIONAL BREAK: ≈0 WHEN THE BREAK IS ABSENT, RISING AND SATURATING NEAR S ≈ 0.4–0.5, WHILE THE MAGNITUDE-BASED AUTOENCODER STAYS AT BASELINE. BANDS: ± 1 S.D. OVER FIVE SEEDS.

Two sweeps bound the regime and close the argument. (a) Holding the estimator fixed and degrading the graph, the gain over the vanilla autoencoder falls from 0.216 at $\rho = 0$ to 0.133 at $\rho = 1$ on the DAG and from 0.168 to 0.025 on the chain: causal structure pays in proportion to graph quality (Fig. 3a). (b) Holding the graph fixed and varying the severity s of the marginal-preserving Type-B break, the gain on Type-B windows rises from ≈0 at $s = 0$ to 0.24 (DAG) and 0.18 (chain) at $s = 1$, saturating near $s \approx 0.4$–0.5, while the graph-free autoencoder stays flat (Fig. 3b). A conditional break is necessary, not merely sufficient: below $s \approx 0.3$ the causal module has no advantage to exploit.

## V. Threats to Validity

All real-benchmark conclusions rest on one model family and its ablations, and the graph estimator is one conditional test. Its two tuning parameters are not load-bearing: over q in {0.01, 0.05, 0.10} no conclusion changes, and over L in {3, 5, 10} recovery is flat except when L falls below the true maximum lag; on the DAG (true $\tau = 5$) halving L to 3 drops edge F1 from 0.997 to 0.487. We separate confirmatory from exploratory analyses: the three factors of Section I are the hypotheses the design tests, whereas the sweeps over $\alpha$, B, $\theta$ and the flagged fraction are exploratory and their p-values are descriptive.

Storage precision. Dumped float32 scores can shift a handful of b = 0 AUCs from the recorded AUROC by up to 0.032 through tie resolution. We therefore re-ran inference in float64 and recomputed VUS-ROC for every entity: across all 24 method comparisons the Holm-corrected decisions do not flip, and the MSL full-versus-vanilla reversal (median −0.029, Holm p = 0.0024) is identical under the two precisions. The largest entity-level |ΔVUS-ROC| is 0.012 (MSL), 0.008 (SMAP) and ≈ 0 on PSM, all far below the 0.032 band, so storage precision is not a threat to our significance claims.

## VI. Conclusion

Holding one causality-guided autoencoder fixed, we varied graph quality, anomaly type and protocol against ground truth. Causal structure earns its complexity where cross-variable structure is genuine and anomalies break conditional relations; on point-adjusted per-channel telemetry it changes nothing measurable. The two deployable tests are graph density (eligibility) and cross-subsample reproducibility (selection inside an eligible system), both measured on the training series alone; the third precondition, that anomalies violate conditional rather than marginal structure, holds everywhere but needs known anomalies to confirm. Enabling the module everywhere costs 0.012 VUS-ROC across the 76 real entities, and the two-stage gate says where that average does not apply.

## References

[1] F. Wang, Y. Jiang, R. Zhang, A. Wei, J. Xie, and X. Pang, "A survey of deep anomaly detection in multivariate time series: Taxonomy, applications, and directions," Sensors, vol. 25, no. 1, art. 190, 2025.

[2] H. Kim, J. Mok, D. Lee, J. Lew, S. Kim, and S. Yoon, "Causality-aware contrastive learning for robust multivariate time-series anomaly detection," in Proc. Int. Conf. on Machine Learning (ICML), 2025. arXiv:2506.03964.

[3] W. Yang, K. Zhang, and S. C. H. Hoi, "A causal approach to detecting multivariate time-series anomalies and root causes," arXiv preprint arXiv:2206.15033, 2022.

[4] X. Wang, Y. Wen, Y. He, H. Xu, Y. Zhao, M. F. C. Haddad, and T. Ma, "CAAD: Causality-aware multivariate time series anomaly detection via multi-scale alignment and structural causal consistency," in Proc. 32nd ACM SIGKDD Conf. on Knowledge Discovery and Data Mining (KDD), vol. 2, pp. 5091-5101, 2026.

[5] D. Cho, J. Han, K. Kang, M. Kim, H. Ryu, and N. Jung, "Structured temporal causality for interpretable multivariate time series anomaly detection," in Advances in Neural Information Processing Systems (NeurIPS), 2025.

[6] S. Kim, K. Choi, H.-S. Choi, B. Lee, and S. Yoon, "Towards a rigorous evaluation of time-series anomaly detection," in Proc. AAAI Conf. on Artificial Intelligence, vol. 36, no. 7, pp. 7194-7201, 2022.

[7] Q. Liu and J. Paparrizos, "The elephant in the room: Towards a reliable time-series anomaly detection benchmark," in Advances in Neural Information Processing Systems (NeurIPS) Datasets and Benchmarks Track, 2024.

[8] P. Boniol, A. K. Krishna, M. Bruel, Q. Liu, M. Huang, T. Palpanas, R. S. Tsay, A. Elmore, M. J. Franklin, and J. Paparrizos, "VUS: Effective and efficient accuracy measures for time-series anomaly detection," The VLDB Journal, vol. 34, no. 3, art. 32, 2025.

[9] J. Audibert, P. Michiardi, F. Guyard, S. Marti, and M. A. Zuluaga, "USAD: Unsupervised anomaly detection on multivariate time series," in Proc. 26th ACM SIGKDD Int. Conf. on Knowledge Discovery and Data Mining (KDD), pp. 3395-3404, 2020.

[10] A. Deng and B. Hooi, "Graph neural network-based anomaly detection in multivariate time series," in Proc. AAAI Conf. on Artificial Intelligence, pp. 4027-4035, 2021.

[11] C. W. J. Granger, "Investigating causal relations by econometric models and cross-spectral methods," Econometrica, vol. 37, no. 3, pp. 424-438, 1969.

[12] K. Hundman, V. Constantinou, C. Laporte, I. Colwell, and T. Soderstrom, "Detecting spacecraft anomalies using LSTMs and nonparametric dynamic thresholding," in Proc. 24th ACM SIGKDD Int. Conf. on Knowledge Discovery and Data Mining (KDD), pp. 387-395, 2018.

[13] A. Abdulaal, Z. Liu, and T. Lancewicki, "Practical approach to asynchronous multivariate time series anomaly detection and localization," in Proc. 27th ACM SIGKDD Int. Conf. on Knowledge Discovery and Data Mining (KDD), pp. 2485-2494, 2021.

[14] F. T. Liu, K. M. Ting, and Z.-H. Zhou, "Isolation forest," in Proc. IEEE Int. Conf. on Data Mining (ICDM), pp. 413-422, 2008.

[15] M. Sakurada and T. Yairi, "Anomaly detection using autoencoders with nonlinear dimensionality reduction," in Proc. MLSDA 2014: 2nd Workshop on Machine Learning for Sensory Data Analysis, ACM, pp. 4-11, 2014.

[16] S. Tuli, G. Casale, and N. R. Jennings, "TranAD: Deep transformer networks for anomaly detection in multivariate time series data," Proc. VLDB Endow., vol. 15, no. 6, pp. 1201–1214, 2022.

[17] Y. Su, Y. Zhao, C. Niu, R. Liu, W. Sun, and D. Pei, "Robust anomaly detection for multivariate time series through stochastic recurrent neural network," in Proc. 25th ACM SIGKDD Int. Conf. on Knowledge Discovery & Data Mining (KDD), 2019, pp. 2828–2837.

[18] J. Xu, H. Wu, J. Wang, and M. Long, "Anomaly transformer: Time series anomaly detection with association discrepancy," in Proc. 12th Int. Conf. on Learning Representations (ICLR), 2022.